\documentclass{article}

\usepackage[preprint]{neurips_2025}

\usepackage{wrapfig}
\usepackage{booktabs}
\usepackage{pifont}
\usepackage[utf8]{inputenc} 
\usepackage[T1]{fontenc}    
\usepackage{hyperref}       
\usepackage{url}            
\usepackage{booktabs}       
\usepackage{amsfonts}       
\usepackage{nicefrac}       
\usepackage{microtype}      
\usepackage{xcolor}         
\usepackage{graphicx}
\usepackage{subcaption}
\usepackage{multirow}
\usepackage{amsmath}

\usepackage{subcaption}
\usepackage{makecell}

\usepackage{hyperref}
\title{HARMONY: Hidden Activation Representations and Model Output-Aware Uncertainty Estimation for Vision-Language Models}

\author{
Erum Mushtaq$^{1}$ \quad  Zalan Fabian$^{1}$ \quad Yavuz Faruk Bakman$^{1}$ \quad Anil Ramakrishna$^{2}$\thanks{This work does not relate to their position at Amazon.}\\ Mahdi Soltanolkotabi$^{1}$ \quad Salman Avestimehr$^{1}$ \\
$^{1}$University of Southern California \quad
$^{2}$Amazon AGI\\
}

\begin{document}

\maketitle

\begin{abstract}
Uncertainty Estimation (UE) plays a central role in quantifying the reliability of model outputs and reducing unsafe generations via selective prediction. In this regard, most existing probability-based UE approaches rely on predefined functions, aggregating token probabilities into a single UE score using heuristics such as length-normalization. However, these methods often fail to capture the complex relationships between generated tokens and struggle to identify biased probabilities often influenced by \textbf{language priors}. Another line of research uses hidden representations of the model and trains simple MLP architectures to predict uncertainty. However, such functions often lose the intricate \textbf{ inter-token dependencies}. While prior works show that hidden representations encode multimodal alignment signals, our work demonstrates that how these signals are processed has a significant impact on the UE performance. To effectively leverage these signals to identify inter-token dependencies, and vision-text alignment, we propose \textbf{HARMONY} (Hidden Activation Representations and Model Output-Aware Uncertainty Estimation for Vision-Language Models), a novel UE framework that integrates generated tokens ('text'), model's uncertainty score at the output ('MaxProb'), and its internal belief on the visual understanding of the image and the generated token (captured by 'hidden representations') at token level via appropriate input mapping design and suitable architecture choice. Our experimental experiments across two open-ended VQA benchmarks (A-OKVQA, and VizWiz) and four state-of-the-art VLMs (LLaVA-7B, LLaVA-13B, InstructBLIP, and Qwen-VL) show that HARMONY consistently matches or surpasses existing approaches, achieving up to 5\% improvement in AUROC and 9\% in PRR.

\end{abstract}

\begin{figure*}[ht]
    \centering
    \includegraphics[width=\linewidth]{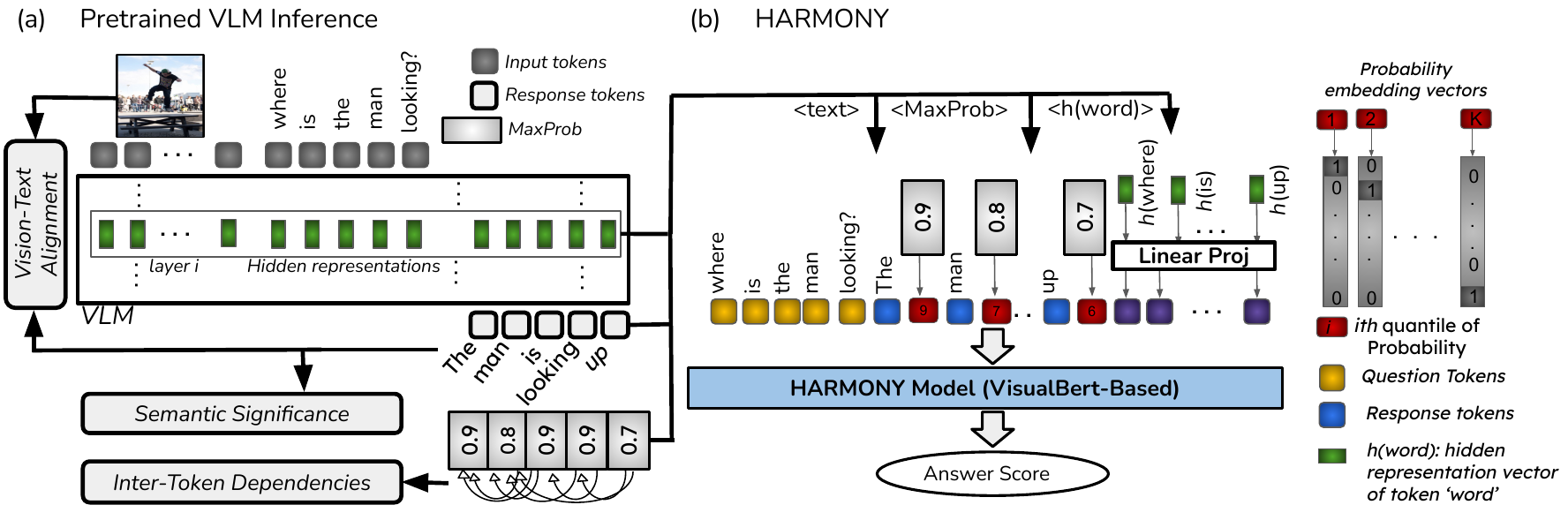} 
    \caption{
        \textbf{Illustration of VLM inference and the proposed framework, HARMONY.} 
        VLMs as autoregressive model take image and question tokens as input and generate text tokens sequentially. During generation, certain tokens (e.g., \textit{``up"} in the example) carry higher semantic relevance to the input, reflecting the inherently complex \emph{inter-token dependencies} in VLM outputs. Moreover, some responses may be weakly grounded in the image and instead dominated by \emph{language priors}, making UE particularly challenging. To address these issues, our proposed framework, \textbf{HARMONY}, jointly leverages the model’s hidden representations as a vision-text alignment signal together with the generated tokens and their sequential token-level probabilities, producing a single UE score for the generated answer.
    }
    \label{fig:two_side_by_side}
\end{figure*}
\section{Introduction}
\label{sec:intro}
\vspace{-7pt}

    

With the rise of Large-Language models, Vision-Language models (VLMs) have seen tremendous growth in recent years. Their remarkable performance in tasks such as visual reasoning, image captioning, and object detection has led to the growing deployment of these models in real-world applications, including assisting the Blind or Low Vision community \cite{ahmetovic2020recog}, medical diagnosis \cite{sepehri2024mediconfusion}, and robotic vision applications \cite{long2024robollm}. However, VLMs often over-rely on language priors \cite{leng2024mitigating} or co-occurring objects \cite{zhou2023analyzing} in an image, which can lead to unreliable or misleading generations. In this regard, Uncertainty Estimation (UE) serves as a valuable approach for evaluating reliability. Unlike other hallucination detection methods that depend on external modules such as object detectors \cite{zhao2024mitigating}, image generations for exploration \cite{augustin2025dash} or image feature extractors \cite{deng2024seeing} to infer \emph{vision entailment}, UE derives reliability from the model’s own internal signals, offering a sophisticated means of assessing model's uncertainty on its generations.

Since VLMs are autoregressive models, they may generate multiple tokens at the output. That is why, some of the key challenges of UE problem for VLMs are 1) capturing semantically significant tokens at the output generation, 2) learning the inter-token dependencies often reflected via the output distribution changes across tokens, and 3) identifying the vision-text alignment at token-level. We illustrate this via an example in Figure \ref{fig:two_side_by_side}, where challenge 1) and 2) would entail identifying the relevant tokens to the given image and question, and learning how the UE distribution changes across tokens to formulate a UE score. Further, challenge 3 requires learning if those tokens are influenced by mere language priors ('model generates tokens based on text priors overlooking the image') or are actually entailed in the image.


Many works have approached the UE problem via black-box formulations \cite{malinin2020uncertainty, kuhn2023semantic,tian2023just, srinivasan2024selective, avestimehr2025detecting}. A key attraction of these methods is that they do not require training, and can work for proprietary-based models. In this regard, some works show that model's consistency can be an indicator of its confidence \cite{khan2024consistency, kuhn2023semantic}. Others argue that self-prompting the model for its own generation can provide a better UE estimate \cite{tian2023just}. Another line of work shows that evidence collection via asking relevant sub-questions can detect unreliable generation if the underlying VLM is well-calibrated, which in itself is a difficult condition to meet \cite{srinivasan2024selective}. 

The other well-known formulation is white-box approach \cite{whitehead2022reliable, dancette2023improving}. This approach requires calibration datasets to train an auxiliary function. In this regard, prior works have shown that the hidden activation representations encode multimodal alignment signal \cite{whitehead2022reliable, dancette2023improving}. They show its effectiveness by leveraging the representations of prompt, and answer to train an MLP-based reliability scoring function. However, such functions often lose the intricate inter-token dependencies. The other related work shows that training a transformer-like architecture on output probabilities can yield a good reliability score \cite{yaldiz2024not}, however, how to encode multimodal alignment signal remains unexplored. 

Given the inter-token dependencies and vision-text alignment challenge, and the findings of prior works \cite{whitehead2022reliable}, \cite{yaldiz2024not}, we hypothesize to leverage model's internal states carrying model's internal understanding of the vision modality, and output probabilities capturing token-level uncertainty such that output distribution shifts and internal belief on multimodal fusion are captured at token level. The key contributions of our works can be summarised as follows:

\begin{itemize}
    \item We propose a novel UE framework, HARMONY (Hidden Activation Representations and Model Output-Aware Uncertainty Estimation for Vision-Language Models), that leverages hidden activations, and output uncertainty scores to predict reliability score of the generated response.
    \item Our experimental evaluations on 2 open-ended VQA benchmarks, 1) AOKVQA, a multimodal reasoning dataset, and 2) VizWiz, visually uncertain/unanswerable questions dataset collected by visually impaired people and four state-of-the-art VLMs (LLaVA-7B, LLaVA-13B, InstructBLIP, and Qwen-VL) show that HARMONY consistently matches or surpasses existing approaches, achieving up to 5\% improvement in AUROC and 9\% in PRR. We also include medical imaging dataset, PathVQA, evaluation comparison.
    \item We also investigate selective prediction performance comparison of the learnable functions, aad further study the effectiveness of HARMONY for out-of-distribution generalization, calibration data sizes, and various input signal designs.
\end{itemize}


\section{Problem Formulation}
\label{problemdef}
\subsection{Uncertainty Estimation}
Though there is no universally accepted definition of UE for LLMs and VLMs \cite{vashurin2025benchmarking}, our work adopts a broadly accepted practical definition from previous works \cite{jiang2024graph, yaldizdesign, huang2024uncertainty}, that is, for a given query \textbf{q}, image $\mathcal{I}$ and generated response $\textbf{s}$, an effective UE should assign a low uncertainty score (indicating higher confidence) if $\textbf{s}$ is \textit{reliable} in the given context. In tasks such as VQA evaluation benchmarks, reliability refers to the correctness of $\textbf{s}$ with respect to the set of ground truth(s) \cite{yaldizdesign}. 

For VLMs, given a question $\textbf{q}$, and an Image $\mathcal{I}$, a model parameterized by $\theta$ generates an output response sequence $\textbf{s} = \{s_{1}, s_{2}, .., s_{k}\}$, where $k$ denotes the length of the sequence. The UE methods quantify the uncertainty for the model's predicted sequence $s$ given the input context. A naive way of estimating uncertainty is to calculate the probability of a generated sequence,
\begin{equation}
    P(\textbf{s}| \textbf{q}, \mathcal{I}, \theta) = \prod_{l=1}^{L} P(s_{l}, |s_{<l}, \textbf{q}, \mathcal{I}, \theta) \label{outputprob}
\end{equation}
where $s_{<l} \overset{\triangle}{=}  \{s_{1}, s_{2}, .., s_{l-1} \}$. It is easy to note that the formulation given in \ref{outputprob} penalizes long sequences. Below, we briefly describe some of the existing probability-based UE methods.

\noindent \textbf{Length-Normalized Scoring} 
\cite{malinin2020uncertainty} fixes the issue of length penalization in sequence probabilities by proposing the following proxy metric,
\begin{equation}
    \tilde{P}(\textbf{s}| \textbf{q}, \mathcal{I}, \theta) = \prod_{l=1}^{L} P(s_{l}|s_{<l}, \textbf{q}, \mathcal{I}, \theta)^{1/L}.
\end{equation}
Their proposed metric essentially normalizes the log probabilities by the length of the sequence.

\textbf{Entropy} \cite{malinin2020uncertainty} is another baseline that leverages Monte-Carlo approximation and beam sampling. It generates multiple beams $B$, and calculates the entropy approximation as 
\begin{equation}
    \mathcal{H}(\theta, \textbf{q}, \mathcal{I}) = -\frac{1}{B}\sum_{b=1}^{B}\text{log}\tilde{P}(\textbf{s}_{b}|\textbf{q}, \mathcal{I}, \theta)
\end{equation}. 

\textbf{Semantic Entropy (SE)} is an improved version of Entropy. It clusters semantically similar generations to reduce the entropy for consistent/semantically similar generations \cite{kuhn2023semantic, farquhar2024detecting}. It sums the scores of all generations belonging to each cluster $\textbf{c}$ as
\begin{equation}
    \tilde{P}(\textbf{c}|\textbf{q}, \mathcal{I}, \theta)=\sum_{\textbf{s} \in c}\tilde{P}(\textbf{s}_{i}|\textbf{q}, \mathcal{I}, \theta)
\end{equation}
and approximates entropy as 
\begin{equation}
    SE(\theta, \textbf{q}, \mathcal{I}) = -\frac{1}{|C|}\text{log}\sum_{i=1}^{C}\tilde{P}(\textbf{c}_{i}|\textbf{q}, \mathcal{I}, \theta)
\end{equation}. 

\textbf{Cluster Entropy} is another variation of Entropy that counts the number of generations in a cluster and calculates the entropy over normalized counts of clusters \cite{kuhn2023semantic}. Note that entropy and SE are computationally expensive methods that require multiple beams for a better estimation of uncertainty. 

\textbf{Self Evaluation} is another popular baseline that asks the model itself to evaluate its own generation and uses the confidence of the correctness token as an uncertainty estimate \cite{tian2023just, srinivasan2024selective}. 

\textbf{First Token} \cite{zhao2024first} is another baseline that addresses the probability aggregation problem by leveraging only the confidence score of first token of the generated response $P(s_{0}, | \textbf{q}, \mathcal{I}, \theta)$ . 

\subsection{Selective Prediction}
A practical use case of uncertainty estimation methods is selective prediction task, where based on the uncertainty estimation function $f(.)$, a decision function $g(.)$ is used to determine whether system choose to answer the question or abstain \cite{el2010foundations}. For the generated sequence $\textbf{s}$ by a VLM, selective system $\mathcal{S}_{\text{VLM}}$ will be as follows,\\
\[
\mathcal{S}_{\text{VLM}}(\textbf{q}, \mathcal{I}) =\begin{cases} 
\textbf{s}, & \text{if } g(\textbf{s})=1 \\
\emptyset, & \text{otherwise }
\end{cases}
\]
where $g(\textbf{s}) = \mathbb{I} \{f(\textbf{s}) > \gamma\}$ given a threshold $\gamma$, $\mathbb{I}$ being an indicator function. Threshold $\gamma$ that provides best differentiation between the correct and incorrect generations is selected from the calibration dataset. $f(.)$ can be any UE function, for example, length-normalized confidence $\tilde{P}(\textbf{s}| \textbf{q}, \mathcal{I}, \theta)$, Entropy $\mathcal{H}(\theta, \textbf{q}, \mathcal{I})$, and Semantic Entropy $SE(\theta, \textbf{q}, \mathcal{I})$ are some of the examples from the above-mentioned UE methods. A model can select to output the prediction if the UE score is above the selected threshold or abstain; output `I don't know' otherwise. In our work, we mainly focus on the use of UE methods that solely rely on the signals from the models, and evaluate them for the selective prediction task.

\section{Related Works}
\label{literaturereview}

The existing uncertainty estimation methods can be broadly categorized into four types: \textit{i) Self-Checking methods, ii) Output Consistency methods, iii) Internal state examination methods and iv) Token Probability methods.}

\textbf{\textit{Self-Checking methods:}} these methods rely on the model's ability to evaluate its own correctness via self-evaluation over its generated answer \cite{tian2023just, srinivasan2024selective}. These works are known for their ability to reduce surface-form competition variations reflected in the output probabilities \cite{holtzman2021surface}, and have been explored for both large-language models (LLMs) \cite{tian2023just} and VLMs \cite{srinivasan2024selective}. However, it has been shown that the self-evaluated confidence of the model is insufficient to be a good estimate of uncertainty \cite{kapoor2024large}.

 \textbf{\textit{Output Consistency methods:}} these methods estimate uncertainty via examining the consistency of the generated output over multiple question rephrasings \cite{farquhar2024detecting, khan2024consistency, shah2019cycle, zhang2024vl} or examining model confidence over relevant sub-questions \cite{srinivasan2024selective}. The question rephrasings \cite{khan2024consistency, zhang2024vl} or beam sampling based methods \cite{farquhar2024detecting} are considered expensive due to multiple forward passes required of the large VLMs. Sub-question-based approaches \cite{srinivasan2024selective} further add to the cost by requiring additional steps such as evidence collection, sub-question formulation, and relevance verification. Additionally, these methods often assume that the VLM is well-calibrated \cite{srinivasan2024selective}, an assumption that may not always hold in practice.
 
\textbf{\textit{Internal state examination methods:}} these works look at the model's hidden activation representations \cite{whitehead2022reliable, dancette2023improving}. They have shown that internal representation of the autoregressive models carry significant reliability signal. They exploit the internal representation vector of input and answer, and train a simple scoring function, such as 2-layer MLP to predict the correctness of the response. While effective, these works require calibration datasets to train the function. 


\textbf{\textit{Token Probability methods:}} these methods use token probabilities at the output to predict the uncertainty by using pre-defined heuristic functions such as length-normalized scoring function \cite{malinin2020uncertainty}, first token probability score \cite{zhao2024first} etc. Some approaches leverage output probabilities and use learnable functions to aggregate probabilities for effective uncertainty estimation \cite{yaldiz2024not} in open-ended question answering. In most cases, VLM-based UE methods frame open-ended visual question answering (VQA) tasks as multiple-choice problems \cite{khan2024consistency, whitehead2022reliable}. Therefore, token probability methods remain relatively unexplored for generative VLMs. 

Our proposed method, HARMONY, integrates both internal state examination and token probabilities, combining their strengths to achieve a more robust UE framework. 


\section{Proposed Method}
\label{HARMONY}

\subsection{Motivation}
\begin{wrapfigure}{r}{0.32\textwidth}
    \vspace{-10pt}
    \centering
    \includegraphics[width=0.32\textwidth]{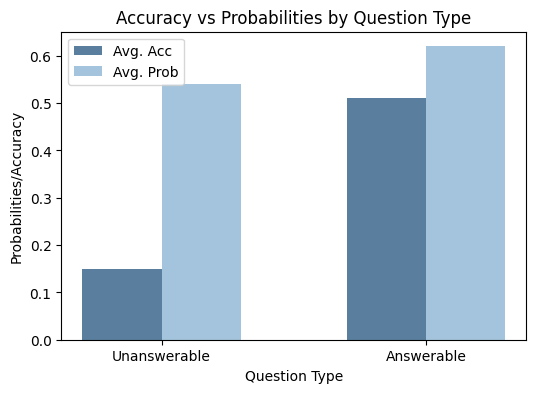}
    \vspace{-10pt}
    \caption{Illustration of language priors challenge where the LLaVa-7B model generates incorrect outputs with high confidence on visually uncertain (unanswerable) questions from the VizWiz dataset.}
    \label{fig2}
    \vspace{-10pt}
\end{wrapfigure}

\textbf{Semantic Significance and Inter-Token Dependencies:}  In free-form generation, VLMs produce multiple tokens in an auto-regressive manner. Estimating uncertainty in this context involves aggregating the probabilities of individual tokens into a single uncertainty estimation score using a predefined scoring function. This makes UE inherently challenging due to factors such as length bias (shorter or longer responses affecting confidence score) \cite{malinin2020uncertainty}, semantic bias (models favoring frequent phrases or syntactic structures) \cite{bakman2024mars} that are often implicit, but significantly impact UE. Various functions proposed in the LLM literature aim to address different aspects of this aggregation process. For example, length-normalized scoring \cite{malinin2020uncertainty} mitigates length bias, while semantic entropy \cite{kuhn2023semantic} captures uncertainty across semantically similar responses. However, identifying an effective aggregation strategy through heuristics remains challenging due to inter-token dependencies and various inherent biases for the given vision and text context. 
\\
\textbf{Language Priors:} For VLMs, assessing token-level semantic significance and inter-token dependencies at the output alone can be insufficient for reliable uncertainty estimation. That is because VLMs are known for their tendency to overlook or misread the evidence in the image, and over-rely on the language-priors as shown in Figure \ref{fig2}. Consider the example question for which the model responds with `The man is looking up'. The uncertainty associated with the token `up' should ideally reflect the model's understanding of the visual scene. However, VLMs may assign a high probability to such tokens due to increasing confidence as the generation progresses, regardless of whether the visual input supports the claim. Therefore, we hypothesize that model's internal belief at token level often captured by internal representations can help address language prior issue. 

\subsection{HARMONY}
To address the inter-token dependencies, and language prior challenge, we aim to design a scoring function that looks at the semantics of the generation ('generated answer': text), token-level uncertainty ('MaxProb':  a real valued number ), and model's internal belief ('internal hidden representations': large dimensional vector) of the model. In addition, we aim the function to learn inter-token dependencies, which requires to preserve the sequential nature of these signals as well. To achieve these objectives, we propose the following function, and input mappings. 

Let $f$ be the scoring function that takes four inputs: the image  conditioned question tokens $\textbf{q} = (q_{1}, q_{2}, .., q_{K})$, the generated response tokens $\textbf{s} =  (s_{1}, s_{2}, .., s_{L})$, the token probabilities $\textbf{p} =  (p_{1}, p_{2}, .., p_{L})$, and the model's hidden-states $\textbf{H} = ({\textbf{h}_{1}, \textbf{h}_{2},..,\textbf{h}_{K}, ..., \textbf{h}_{K+L}})$ corresponding to $\textbf{q}$ and $\textbf{s}$. Here, $p_{i}$ denotes the probability of generated sequence token $i$, and each hidden state 
$\textbf{h}_{i} \in R^N$ vector, where N represents the number of hidden units at a specific layer.  It is worth-noting that the first $K$ vectors in $\textbf{H}$ correspond to the question tokens, while the remaining $L$ represent the generated tokens. We use the hidden states of the tokens right after the visual tokens, as they inherently encode cross-modal fusion signal, capturing information transferred from visual tokens to text tokens. Note that our inputs consist of varied nature i.e, texts, probabilities which are real numbers, and hidden states which are high dimensional vectors requiring a structured approach to model their inter-dependencies in a sequential manner.

\textbf{Scoring Function} Given the sequential nature of input data, we leverage the pretrained VisualBERT architecture, an extension of the encoder-based BERT model. VisualBERT is by design suitable for taking text and high-dimensional visual features as inputs. It maintains two separate sets of embeddings $E$ and $F$ that correspond to text and visual features, respectively. For text data, it tokenizes the input and maps each token to a set of embeddings, $e \in E$. Likewise, for the vision context, it leverages pre-computed high-dimensional visual features corresponding to different regions of the images. It takes the visual features as input and assigns them an embedding $f \in F$. 

\textbf{Input Mapping} VisualBERT is naturally well-suited for textual input. Therefore, question and answer text tokens can be easily integrated into the input to the model. However, for hidden states, we project hidden representations to the space of model's visual embeddings via linear projection, and use them as an input. Further, to encode probability information, we leverage a third set of embedding which is inspired by work \cite{yaldiz2024not}. The key idea is that the probability range [0,1] can be split into a fixed k partitions. For the given dimension $d$ of input embedding, if $p_i$ falls in the range of r-th partition, the vector positions between $(r-1) \times kd$ and $r \times kd$ are set to one while all other positions are set to zero. This allows representation of distinct probability ranges via orthogonal embedding vectors. 

\textbf{Learnable Task} At the input, we have question, followed by generated tokens and their corresponding probabilities, further followed by hidden representations of the question and the generated tokens as shown in the Figure \ref{fig:two_side_by_side}. Given these input, the task is to predict the correctness of the generation $\textbf{s}$ for which we augment the VisualBERT model at the output via linear layer that gives a single logit output, $f(\textbf{q}, \textbf{s}, \textbf{H}, \textbf{p})$. We employ binary cross-entropy loss, 
\begin{align}
    \mathcal{L}\big(f(\mathbf{q}, \mathbf{s}, \mathbf{H}, \mathbf{p}), g\big) 
    &= -  
        g \log f(\mathbf{q}, \mathbf{s}, \mathbf{H}, \mathbf{p}) \\&
        + (1 - g) \log\big(1 - f(\mathbf{q}, \mathbf{s}, \mathbf{H}, \mathbf{p})\big)
    \label{eq:harmony_loss}
\end{align}

where $g$ is the binary ground-truth label (of accuracy for generation $\textbf{s}$ as an answer to $\textbf{q}$) used here as a target.  Note that both VisualBERT model and the linear projection layer are fine-tuned on this reliability score prediction task. It is worth mentioning that our proposed framework contributes at scoring function level, which can be integrated into other mixture-of-experts setup based baselines \cite{dancette2023improving}. 

\section{Results}
\label{results}
\subsection{Experimental Setup}
\textbf{Datasets and Models} We evaluate our work on two VQA datasets, A-OKVQA \cite{schwenk2022okvqa} and VizWiz \cite{gurari2018vizwiz}. The A-OKVQA datasets require reasoning and common sense alongside visual information. The VizWiz dataset covers a challenging setup, where each image is taken by a blind/visually impaired individual and accompanied by spoken questions about the images. These questions are then transcribed. We also include experiments on PathVQA \cite{he2020pathvqa} which is a medical imaging VQA dataset. To train the scoring function, we leverage the training splits of the corresponding datasets. We use a 80\% and 20\% split to construct a train and a validation split, respectively. To evaluate the performance of the scoring function, we use the validation split given with the dataset as a test split. We provide further details of datasets and training strategy/hyper-parameters in Appendix \ref{trs}. We evaluate our method on four open-sourced VLMs: LLaVA-7b \cite{liu2023visual}, LLaVA-13b \cite{liu2024improved}, InstructBLIP \cite{huang2023visual} and Qwen-VL \cite{bai2023qwen}. InstructBLIP uses FlanT5-XL \cite{raffel2020exploring} as the LLM backbone. All these models are instruction-tuned on the VQA task. 
\begin{table*}[!ht]
\caption{AUROC and PRR scores on A-OKVQA and VizWiz dataset }
\vspace{-6pt}
\centering
\fontsize{6}{5}\selectfont
\resizebox{1\textwidth}{!}{
\begin{tabular}{ cl c c c c c c c c}
\toprule
& & \multicolumn{2}{c}{\textbf{LLaVA - 7b}} & \multicolumn{2}{c}{\textbf{LLaVA - 13b}} & \multicolumn{2}{c}{\textbf{Instruct-BLIP}} & \multicolumn{2}{c}{\textbf{Qwen-VL}} \\
 
 \midrule
 
& UE Method & AUROC(\%)  & PRR(\%) & AUROC(\%)  & PRR(\%)  & AUROC(\%) & PRR(\%)  & AUROC(\%) & PRR(\%)\\
\midrule[\heavyrulewidth]
\multirow{13}{*}{\rotatebox{90}{AOKVQA}}
&Length-Normalized Confidence     & 74.55 & 61.45     & 77.50   & 67.04 & 74.13 & 56.60 &65.53&33.83\\
&First Token Confidence   & 69.39 &33.16& 72.96 & 41.35 & 75.09 & 58.47 &63.59&28.28 \\
&Self-Eval Confidence         & 71.53 & 54.48 & 63.04 & 54.43& 76.12 & 62.75 &58.06&10.81\\
\cmidrule{2-10}
&Entropy       & 61.38 & 35.65 & 67.57 & 49.23& 54.15& 30.15 &69.11&48.48 \\
&Semantic Entropy      & 78.39 & 68.48 & 80.83 & 69.89& 73.72 &   52.20 &59.05&11.05\\
&Cluster Entropy      & 69.87 &  52.27 & 68.90 &50.89& 71.00 &51.11 &59.05&18.08 \\
\cmidrule{2-10}
& MSF &78.66 & 67.01&77.64&67.07&79.93&67.31&76.82&64.94\\
&LARS & 79.90 &68.95&81.46&73.70&80.07&68.31&74.63&54.37\\
&HARMONY [Ours] & \textbf{83.99} & \textbf{75.05} &\textbf{83.72} & \textbf{77.09}&\textbf{81.73}&\textbf{72.03}&\textbf{81.84}&\textbf{74.10}\\
&&\textbf{(+4.09)}&\textbf{(+6.10)}&\textbf{(+2.26)}&\textbf{(+3.36)}&\textbf{(+1.66)}&\textbf{(+3.72)}&\textbf{(+5.02)}&\textbf{(+9.14)}\\
\midrule[\heavyrulewidth]

\multirow{13}{*}{\rotatebox{90}{VizWiz}}

&Length-Normalized Confidence  &  71.94&44.30 &75.61&55.32&77.57&75.51& 71.97&45.10\\
&First Token Confidence   & 69.76&43.06&71.52&46.65&76.00&56.48&70.88&39.20\\
&Self-Eval Confidence         & 63.09&30.62&67.22&42.06&73.43&52.01&64.13&27.12\\
\cmidrule{2-10}
&Entropy       & 33.84&15.16&41.54&19.32&38.31&28.72&56.71&16.15\\
&Semantic Entropy      & 64.04&21.82&70.89&38.21&66.43&27.90&66.60&26.02\\
&Cluster Entropy      & 59.86&18.83&65.23&30.25&66.09&30.40&66.99&27.03\\
\cmidrule{2-10}
& MSF  &85.66&74.43&86.43&74.42&86.62&73.67&85.10&71.47\\
&LARS & 80.50 & 64.13&85.29&73.09 &86.34&72.14&81.37&62.94\\
&HARMONY [Ours] & \textbf{87.26} & \textbf{76.83}&\textbf{88.71}&\textbf{79.68}&\textbf{86.63}&\textbf{73.75}&\textbf{87.47}&\textbf{76.89}\\
&&\textbf{(+1.60)}&\textbf{(+2.40)}&\textbf{(+2.28)}&\textbf{(+5.26)}&\textbf{(+0.01)}&\textbf{(+0.08)}&\textbf{(+2.37)}&\textbf{(+5.42)}\\

\bottomrule
\end{tabular}
}
\label{UE_comparison_}
\end{table*}

\textbf{Evaluating UE Performance}
To assess the correctness of generated outputs, we employ $\text{LAVE}_{\text{GPT-3.5}}$ \cite{manas2024improving} as an evaluator, following the approach of prior work \cite{srinivasan2024selective}. LAVE employs a large language model to estimate the semantic similarity of each predicted answer to the crowdsourced answers in the benchmark. We regard a score greater than 0 (one or more matches) as correct label and a score of 0 (no-matches) as incorrect. Following previous UE works on auto-regressive models \cite{bakman2024mars, yaldiz2024not}, we use AUROC (Area Under the Receiver Operating Characteristic) as our evaluation metric. It is commonly used to evaluate the performance of binary classifiers \cite{kuhn2023semantic}. The score range for AUROC is 0.5 (random) to 1 (perfect).
We also report the prediction rejection ratio (PRR), another widely used metric for evaluating UE in \cite{malinin2020uncertainty}. PRR quantifies the relative precision gain obtained by rejecting low-confidence predictions, measuring how much precision improves as increasingly uncertain outputs are discarded \cite{UEsurvey}. It can be defined as the gap between the area under the rejection curve (AUC) of the evaluated uncertainty scores and that of a random baseline, normalized by the gap between an oracle UE baseline and the same random baseline:
\begin{equation}
\text{PRR} = \frac{\text{AUC}_{\text{baseline}} - \text{AUC}_{\text{rand}}}{\text{AUC}_{\text{oracle}} - \text{AUC}_{\text{rand}}}
\end{equation}
where $\text{AUC}_{\text{baseline}}$ signify the area under the precision-rejection curve for the given baseline method, $\text{AUC}_{\text{oracle}}$ are the oracle scores aligning perfectly with the correctness, and $\text{AUC}_{\text{rand}}$ corresponds to the random rejection. The PRR ranges from 0 (random) to 1 (perfect). 

\textbf{Evaluating Selective Prediction Performance}
Following the previous work \cite{whitehead2022reliable}, we also evaluate the performance of our scoring function on threshold based evaluation by computing the coverage and risk, and effective reliability (ER) metrics which are explained below.

\textit{\textbf{Coverage, Risk and ER}} Coverage is the portion of questions that model opted to answer. That is, given the decision function $g(.)$ on the dataset $\mathcal{D}$ with input $\textbf{s}_{i}$, coverage is defined as:
\begin{equation}
\mathcal{C}(g) = \frac{1}{|D|} \sum_{\textbf{s}_{i} \in D} g(\textbf{s}_{i})
\end{equation}
whereas risk is the error on the portion of questions covered by the model such as:
\begin{equation}
\mathcal{R}(g) = \frac{\sum_{\textbf{s}_{i} \in D} (1- \text{Acc}(\textbf{s}_{i})).g(\textbf{s}_{i})}{\sum_{\textbf{s}_{i} \in D} g(\textbf{s}_{i})}
\end{equation}
where Acc(.) is the accuracy of the generated sequence $\textbf{s}_{i}$. Note from the definition of $g(.)$ that for lower thresholds, model covers more questions, however, risk on those questions increases. Therefore, at different risk levels, we obtain different coverages. In our evaluation, we evaluate coverage at $10\%$ and $20\%$ risk levels. An ideal UE estimate should yield low-risk and high coverage. ER calculates these two characteristics by assigning a reward of 1 to each question that is answered correctly, penalizes the questions that are answered wrong by a cost of 1, and gives zero reward to the questions that model abstains on. To calculate ER, we compute the threshold maximizing ER on the validation split of the calibration set, and use that threshold on the test set to report the performance. 

\textbf{Baselines}
We include a range of black-box methods, including length-normalized confidence \cite{malinin2020uncertainty}, first-token confidence \cite{zhao2024first}, and self-eval \cite{srinivasan2024selective}, Entropy, Semantic Entropy, and Cluster Entropy \cite{farquhar2024detecting}. Semantic Entropy, which measures consistency among semantically similar answers and can be regarded as an alternative implementation of \cite{khan2024consistency}. Finally, we consider supervised training-based methods, including MSF (Multimodal Selection Function) \cite{whitehead2022reliable}, which trains an MLP on hidden representations of the prompt and generated answer, and LARS \cite{yaldiz2024not}, which trains a transformer architecture on the token probabilities predicted by the base model. Details of black-box methods can be found in section \ref{problemdef} and white-box methods in Appendix \ref{baseinfo}.

\subsection{Results}

\subsubsection{UE Performance}

\begin{table*}[ht]
\caption{ Selective Prediction Performance: Coverage at risks (10\% \& 20\%) and ER (cost=1)}
\vspace{-6pt}
\centering
\fontsize{20}{12}\selectfont
\renewcommand{\arraystretch}{2.4}
\resizebox{0.9\textwidth}{!}{
\begin{tabular}{ cl c c c  |c c c  |c c c | c c c}
\toprule
& & \multicolumn{3}{c}{\textbf{LLaVa - 7b}} & \multicolumn{3}{c}{\textbf{LLaVa - 13b}} & \multicolumn{3}{c}{\textbf{Instruct-BLIP}} & \multicolumn{3}{c}{\textbf{Qwen-VL}} \\
 
 \midrule
 
& UE Method & ER(\%) & C@R=10\% & C@R=20\%&  ER(\%) & C@R=10\% & C@R=20\%& ER(\%)  & C@R=10\% & C@R=20\% &ER(\%)  & C@R=10\% & C@R=20\%\\
\midrule[\heavyrulewidth]
\multirow{3}{*}{\rotatebox{90}{AOKVQA}}
& MSF &49.17&43.75&80.17&53.19&50.56&89.52&38.15&32.66&60.08&53.62&46.73&87.77\\
&LARS &  49.78&50.65&82.53&53.71 &61.83 & 90.48&37.73&31.27&61.83&53.72&43.14&90.66\\
&HARMONY [Ours]  & \textbf{52.31}& \textbf{60.61}&\textbf{85.59} & \textbf{55.90}&\textbf{64.80}&\textbf{92.23}&\textbf{38.25}&\textbf{36.77}&\textbf{63.32}& 55.63&59.65&90.74\\
\midrule[\heavyrulewidth]
\midrule[\heavyrulewidth]
\multirow{3}{*}{\rotatebox{90}{VizWiz}}

& MSF &21.30&15.00&34.15&23.47&11.90&37.76&13.01&9.15&19.84&16.76&6.01&22.57\\
&LARS & 15.72 & 4.91 & 21.53&19.14&8.86&29.08&12.82&6.85&17.13&13.09&1.05&15.34\\
&HARMONY [Ours] & \textbf{21.93}& \textbf{16.37}&\textbf{36.03}&\textbf{24.69}&\textbf{20.24}&\textbf{41.11}&\textbf{13.38}&\textbf{9.23}&\textbf{19.94}&\textbf{18.08}&\textbf{9.64}&\textbf{28.27}\\
\bottomrule
\end{tabular}
}
\label{select_pred}
\end{table*}

We present the comparison results of HARMONY against existing UE baselines on the A-OKVQA and VizWiz datasets in Table~\ref{UE_comparison_}. We observe that learnable functions such as MSF, LARS, and HARMONY generally perform better across considered datasets and models. Although these methods require a one-time function training cost, they are often more efficient than some of the black-box approaches (cluster entropy, entropy, semantic entropy), which typically require multiple forward passes of the underlying 7B or 13B models. In contrast, HARMONY consists of only 113M trainable parameters and requires a single forward pass to compute the reliability score. Between MSF and LARS, we find that on the VizWiz dataset which contains visually unanswerable questions, MSF surpasses LARS, highlighting the importance of leveraging hidden states information as a multimodal alignment signal in visually uncertain settings. However, for InstructBLIP model with  VizWiz dataset, we find that all considered white-box method converge to similar performance levels. Besides that, HARMONY consistently outperforms both LARS and MSF, achieving up to a 5\% improvement in AUROC and a 9\% improvement in PRR across rest of the models and datasets.

\begin{wraptable}{r}{0.51\textwidth}
\vspace{-16pt}
\caption{UE Performance on PathVQA}

\centering
\renewcommand{\arraystretch}{1.0}
\setlength{\tabcolsep}{9pt}
\resizebox{0.5\textwidth}{!}{
\begin{tabular}{ cl c c }
\toprule
& & \multicolumn{2}{c}{\textbf{LLaVA - 13b}}   \\
 
 \midrule
& UE Method & AUROC(\%)  & PRR(\%)  \\
\midrule[\heavyrulewidth]
&Length-Normalized Confidence  &  82.35&55.59 \\
&First Token Confidence   & 82.27&54.35\\
&Self-Eval Confidence         & 63.81&35.50\\
\cmidrule{2-4}
&Entropy       & 70.71&32.88\\
&Semantic Entropy      & 64.15&36.27\\
&Cluster Entropy      & 64.97&35.93\\
\cmidrule{2-4}
& MSF  &96.53&93.07\\
&LARS & 96.89 & 93.69\\
&HARMONY [Ours] & \textbf{97.31} & \textbf{94.80}\\
&&\textbf{(+0.42)}&\textbf{(+1.14)}\\

\midrule[\heavyrulewidth]

\end{tabular}
}
\vspace{-20pt}
\label{medical}
\end{wraptable}

\textbf{Medical imaging Dataset:} We also report the UE performance on medical imaging domain dataset, PathVQA. For this dataset, we find that the output probability based scoring functions perform significantly lower than the learnable functions as shown in \ref{medical}.  Further, HARMONY achieves state-of-the-art performance indicating its effective performance in medical domain as well.

\subsubsection{Selective Prediction Performance}
Here, we present the comparison of the learnable scoring functions baselines on the selective prediction task. Before we compare these methods, it is important to mention that for all trainable functions, we select the best model checkpoints based on the AUROC scores. However, a user may choose a different criterion such as ER to achieve higher performance on this downstream task. The objective here is to present a practical use case, and compare the performance of these methods. 
\begin{figure*}[ht]
    \centering
    \includegraphics[width=0.7\linewidth]{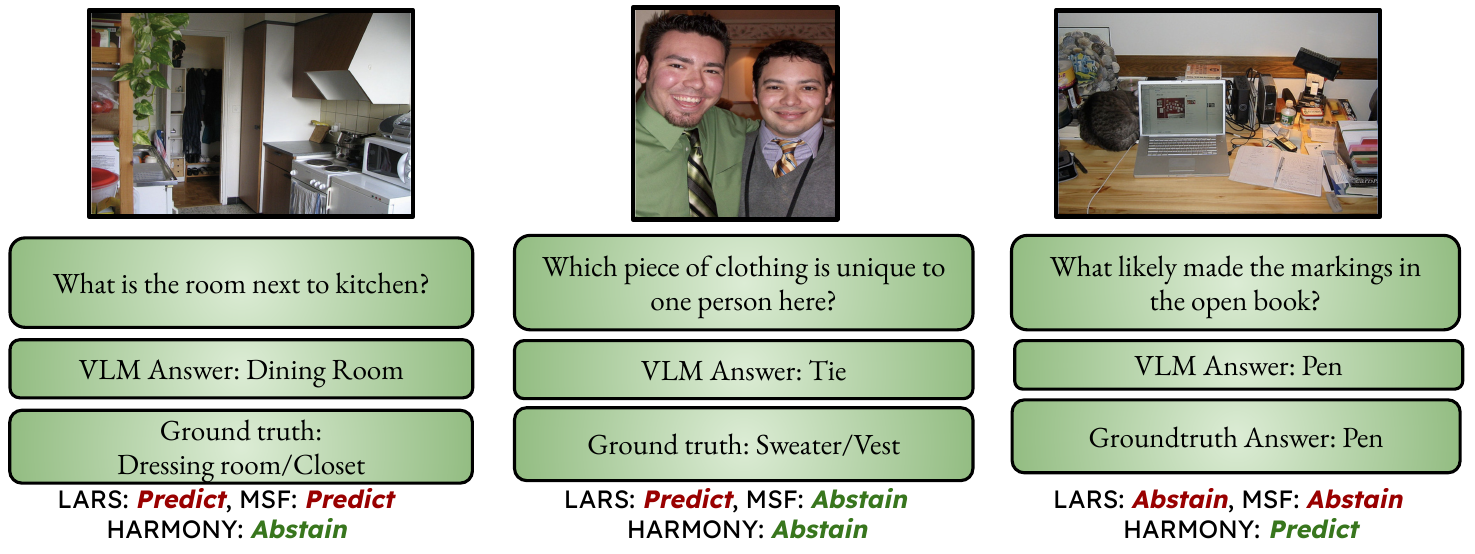}
    \caption{An illustration of selective prediction decisions on the A-OKVQA dataset with LLaVa-7b model. In the left-most example, the model generates an incorrect answer, yet both LARS \cite{yaldiz2024not} MSF \cite{whitehead2022reliable} choose to answer/predict based on their respective calibration thresholds. In the second example, LARS opts to answer, while MSF correctly abstains. In the right-most example, both methods abstain, whereas HARMONY makes the right prediction for each of these examples.}
    \label{fig:abstention_example}
\end{figure*}
First, we report how many questions are covered by our method at 10\% and 20\% risk levels. The larger the number, the better performance. We observe that our proposed method consistently achieves similar or higher coverage across various models, and datasets. We also report ER metric, which represents a better tradeoff between coverage and risk due to a penalty on the incorrectly covered question. For the comparison, we select a threshold for each method giving best ER on the validation split of calibration set, and compute effective reliability using that threshold on the test set. For this metric as well, our method either performs similar or outperforms other methods achieving up to 3.14\% higher score. This highlights its potential to yield higher coverage while inuring lower risks. As an example, we present some sample questions in Figure \ref{fig:abstention_example} and compare the decision predictions on the trainable functions. While training on either output distributions or hidden representations alone can lead to contradictory or consistently incorrect decisions, leveraging both simultaneously results in better decision functions.



\begin{wrapfigure}{r}{0.45\textwidth}
\centering
\includegraphics[width=\linewidth]{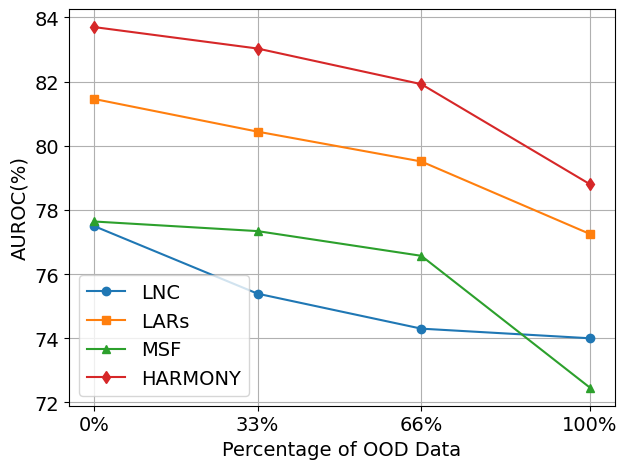}
\captionsetup{font=footnotesize}
\caption{Out-of-Distribution Generalization.}
\label{fig:1c}
\vspace{-20pt}
\end{wrapfigure}

\subsubsection{Out-of-Distribution Generalization}
To evaluate out-of-distribution (OOD) generalization, we test the LLaVA-13B model trained on the A-OKVQA dataset, which focuses on visual reasoning, by introducing OOD samples from the OKVQA dataset \cite{marino2019ok}. The experimental setup, illustrated in Figure \ref{fig:1c}, progressively increases the proportion of OOD samples in the evaluation set. Specifically, the x-axis denotes the percentage of OOD data, where, for example, 33\% corresponds to a test mixture containing 33\% OKVQA (OOD) samples and 66\% A-OKVQA (in-distribution) samples. As the proportion of OOD data increases, we observe a consistent decline in AUROC scores across all supervised baselines, including model confidence based scoring such as Length Normalized Confidence indicating a degradation in uncertainty calibration under distributional shift. However, HARMONY comparatively achieves higher AUROC values, demonstrating relatively better generalization to unseen data distributions.
\subsubsection{Ablation across model layers and input signals}


\begin{wraptable}{r}{0.40\textwidth}
\vspace{-30pt}
\centering
\caption{\small AUROC (\%) of HARMONY and MSF on LLaVA-7B across hidden layers. }
\scriptsize
\renewcommand{\arraystretch}{1.0}
\setlength{\tabcolsep}{15pt}
\begin{tabular}{ccc}
\toprule
\textbf{Layer} & \textbf{HARMONY} & \textbf{MSF} \\
\midrule
32 & 80.97 & 75.92 \\
24 & 81.77 & 76.43 \\
20 & 82.84 & 76.61 \\
16 & \textbf{83.99} & 77.76 \\
12 & 82.69 & \textbf{78.66} \\
8  & 80.83 & 76.75 \\
\bottomrule
\end{tabular}
\label{tab:ablation_layers_short}
\vspace{-8pt}
\end{wraptable}

Our hypothesis is that internal layers have a signal of multimodal reliability, however, it is not clear which layer would provide the best signal. Previous works have highlighted that inner layers (layers closer to the input) focus more on extracting lower-level information from the input, while outer layers (layers closer to the model output) are mostly focused on the next token generation \cite{azaria2023internal}. Therefore, for all the models, we ablate over every fourth layer for both MSF, and our method. We report the best performing layer results in Table \ref{UE_comparison_} for LLaVa-7b. We observe that for LLaVa-7b and 13b models, and Qwen-VL, inner layers (layer 16 and layer 22) yield the best AUROC performance for our method across both datasets. Further, for InstructBLIP, we find the outer-most layer performs the best.

\begin{wraptable}{r}{0.5\textwidth}
\centering
\captionsetup{font=footnotesize}
\caption{Ablation over input signals, generated tokens (text), probability associated with each token (Prob), and hidden states of the input prompt, on LLaVA-13B model and A-OKVQA dataset.}
\scriptsize
\renewcommand{\arraystretch}{0.9}
\setlength{\tabcolsep}{3pt}
\begin{tabular}{lcccccc}
\toprule
\textbf{Architecture} & \textbf{Text} & \textbf{Prob} & \textbf{HS} & \textbf{PRR (\%)} & \textbf{AUROC (\%)} \\
\midrule
VisualBERT & \checkmark & $\times$ & $\times$ & 23.06 & 59.45 \\
VisualBERT & \checkmark & \checkmark & $\times$ & 72.17 & 80.23 \\
VisualBERT & \checkmark & $\times$ & \checkmark & 71.98 & 80.72 \\
\textbf{VisualBERT (Ours)} & \checkmark & \checkmark & \checkmark & \textbf{77.09} & \textbf{83.72} \\
\bottomrule
\end{tabular}
\label{tab:partial_information}
\end{wraptable}

We also conduct an ablation study on partial input signals, generated tokens (Text), token-level probabilities (Prob), and hidden states (HS), using the LLaVA-13B model on the A-OKVQA dataset. We use VisualBERT model as the scoring function transformer architecture for finetuning for the sake of consistency. As shown in \ref{tab:partial_information}, we find that text (question and generated answer) without the token probabilities and hidden states yield no significant UE estimate. However, addition of token probabilities accompanying each token text helps the scoring function learn a better UE estimate. Likewise, if we use text and hidden states of the generated tokens as input signals, they yield better results than the MLP based MSF function. Further, combining all three sources of information as proposed in our work yield significantly better reliability estimates. 
\vspace{20pt}

\begin{wrapfigure}{r}{0.45\textwidth}
\centering
\includegraphics[width=\linewidth]{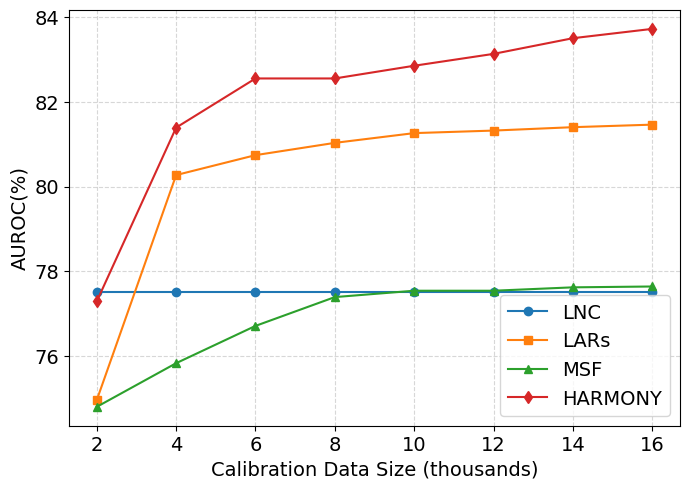}
\captionsetup{font=footnotesize}
\caption{Effect of calibration data size on the UE performance.}
\label{fig:1b}
\end{wrapfigure}

\subsubsection{Effect of Calibration Data}
Next, we conduct an ablation study to investigate how the function's performance scales with the amount of calibration data. For this, we vary the calibration set size with LLaVA-13b model on AOKVQA dataset and report AUROC. As shown in Figure \ref{fig:1b}, increasing the calibration data size consistently enhances the AUROC scores across all learnable baselines. We also observe that HARMONY requires at least 2,000 samples to achieve performance comparable to LNC, and approximately 6,000 samples to surpass existing SOTA baselines. This trend highlights that while learnable approaches benefit significantly from more calibration data, their relative advantage becomes more apparent only once sufficient data are available to capture the diversity of uncertainty patterns in multimodal settings.

\vspace{-2pt}
\section*{Limitations} HARMONY requires labeled calibration data to train the scoring function. While it is needed for other learnable scoring functions (LARS, MSF) as well, the black-box method may not have this limitation. Further, it relies on access to the model’s internal representations, which makes it incompatible with closed-source models. HARMONY may also require architecture-specific adjustments such as for the linear layer projection component, as internal hidden state dimensions vary across VLM families. Addressing these constraints is an important direction for future research on developing reliable and generalizable VLM UE methods.


\vspace{-9pt}
\section*{Conclusion}
This work introduces a novel UE method HARMONY for VLMs that effectively combines hidden activation representations with output token probabilities and generated token text. By jointly leveraging model internal states and output beliefs at token level, our proposed framework provides a more holistic reliability assessment, complementing probability-based and representation-based approaches. Our extensive experiments on AOKVQA and VizWiz datasets demonstrate that our method consistently matches or surpasses existing approaches, achieving up to 5\% improvement in AUROC and 9\% in PRR.

\bibliographystyle{plain}
\bibliography{neurips_2025}

\newpage

\newpage
\appendix

\section{Appendix}
\label{addon}






\subsection{Learnable Scoring Function Baselines}
In our work, we compare our method, HARMONY, with two existing state-of-the-art baselines, multimodal selection function (MSF) \cite{whitehead2022reliable} and LARS \cite{yaldizdesign}. In the following, we explain these two learnable scoring baselines.
\label{baseinfo}
\subsubsection{Multimodal Selection Function (MSF)} 
This work demonstrates the importance of hidden representations in capturing uncertainty in model generation. The key premise of MSF is that using hidden representations of the image, question and answer as input to a simple two-layer MLP can effectively estimate the uncertainty \cite{whitehead2022reliable}. These hidden representations at a given layer have shape [context length,vector dimension] and MSF aggregates them using max or mean pooling across the context length dimension.
In autoregressive VLMs, the question tokens usually follow the image, so the hidden representations of the question tokens already encode the relevant visual information through the causal attention mechanism. Furthermore, the large-scale models often contain a substantial number of image tokens (e.g., 576 in LLaVA and 184 in Qwen-VL), many of which are redundant with information already captured in the question-token representations due to image conditioning/prior context. To show this, we performed ablation studies on LLaVA in the AOKVQA and VizWiz datasets and find that the addition of image token hidden representations, reported as a baseline $\text{MSF}^{*}$ in Table \ref{tab:llava_auroc_prr} does not yield improvement over MSF (without image token hidden representations). To ensure a fair comparison, we, therefore, use the question-token and answer-token representations as input to both the two-layer MLP and our method for all results in Table~\ref{UE_comparison_}. Under this set of inputs, we show that our method more effectively captures uncertainty by exploiting variation across the context-length dimension, monitoring token-level internal belief signals from hidden representations and output uncertainty from probability embeddings.

\begin{table}[!ht]
\centering
\caption{MSF and $\text{MSF}^{*}$ comparison for LLaVA-7B and LLaVA-13B on AOKVQA and VizWiz datasets.}
\vspace{-3pt}
\scriptsize
\begin{tabular}{l l c c}
\toprule
\textbf{Dataset} & \textbf{UE Method} & \textbf{LLaVA-7B} & \textbf{LLaVA-13B} \\
 &  & AUROC / PRR (\%) & AUROC / PRR (\%) \\
\midrule
AOKVQA
 & MSF        & 78.66 / 67.01 & 77.64/67.07\\
 & $\text{MSF}^{*}$       & 78.42 / 66.99 & 77.61/ 67.01 \\
 & HARMONY    & \textbf{83.99 / 75.05} & \textbf{83.72 / 77.09} \\
\midrule
VizWiz
 & MSF        & 85.66 / 74.43 & 86.43 / 74.42 \\
 &$\text{MSF}^{*}$       & 85.42 / 73.14 & 86.15 / 74.38 \\
 & HARMONY    & \textbf{87.26 / 76.83} & \textbf{88.71 / 79.68} \\
\bottomrule
\end{tabular}
\label{tab:llava_auroc_prr}
\end{table}

\subsubsection{LARS}
LARS is a state-of-the-art baseline that captures uncertainty in generation by leveraging token-level uncertainty scores at the output \cite{yaldizdesign}. The method trains a BERT transformer architecture and achieves competitive performance, outperforming various existing black-box approaches. Our work advances this line of research by combining hidden representations with token-level uncertainties, leading to significant performance gains over LARS.

\subsection{Training Setup}
\label{trs}
\subsubsection{Dataset Information}
\begin{figure*}[!h]
    \centering

    \begin{subfigure}[t]{0.25\textwidth}
        \centering
        \includegraphics[width=\linewidth]{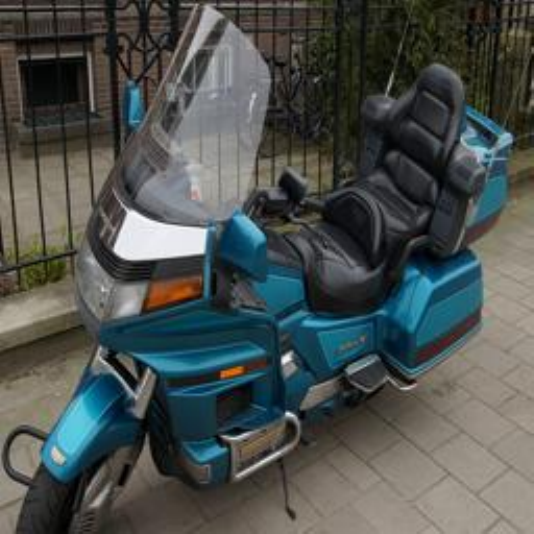}
        \vspace{3pt}
        \textbf{Q:} How many people can ride this motorcycle at a time?\\
        \textbf{A:} Two
        \caption{AOKVQA sample}
        \label{fig:aokvqa}
    \end{subfigure}
    \hspace{0.015\textwidth}
    \begin{subfigure}[t]{0.25\textwidth}
        \centering
        \includegraphics[width=\linewidth]{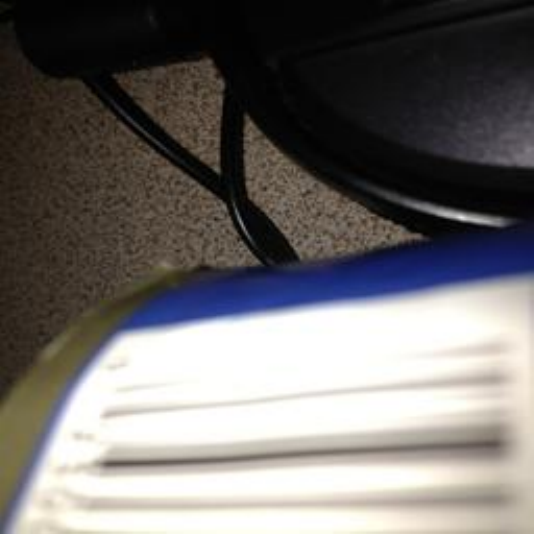}
        \vspace{3pt}
        \textbf{Q:} What can is this?\\
        \textbf{A:} Unanswerable
        \vspace{10pt}
        \caption{VizWiz sample}
        \label{fig:vizwiz}
    \end{subfigure}
    \hspace{0.015\textwidth}
    \begin{subfigure}[t]{0.25\textwidth}
        \centering
        \includegraphics[width=\linewidth]{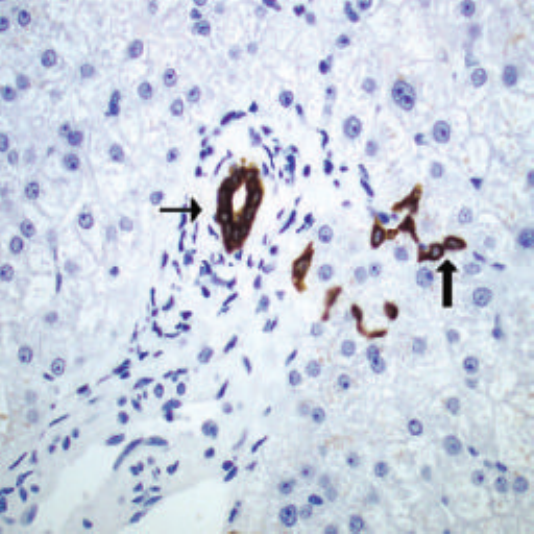}
        \vspace{3pt}
        \textbf{Q:} Where are liver stem cells (oval cells) located?\\
        \textbf{A:} In the canals of Hering
        \caption{PathVQA sample}
        \label{fig:pathvqa}
    \end{subfigure}

    \caption{Example question and answer pairs from the three datasets used in our experiments: 
    AOKVQA (subfigure~\ref{fig:aokvqa}), VizWiz (subfigure~\ref{fig:vizwiz}), and PathVQA (subfigure~\ref{fig:pathvqa}).}
    \label{fig:example_samples}
\end{figure*}

We perform evaluations on three datasets: AOKVQA, an image reasoning dataset; VizWiz, a visually answerable or unanswerable image dataset; and PathVQA, a medical imaging dataset. Figure~\ref{fig:example_samples} provides example image and question–answer pairs from these datasets to demonstrate the diverse range. For each dataset, we use the provided training split as a calibration dataset. The calibration data is further divided into 80\% training and 20\% validation partitions to train our scoring function. To evaluate our method, we used the original validation split given with the dataset as the test set. This calibration and test split setup is applied consistently across all three datasets. Details of the resulting train/validation/test splits are provided in Table~\ref{tab:calib}.

\begin{table}[!h]
\centering
\small
\setlength{\tabcolsep}{6pt} 
\caption{Calibration and Test Data Sizes}
\begin{tabular}{ccc}
\toprule
\textbf{Dataset Name} & Calibration Data (\textbf{Train} + \textbf{Val}) & \textbf{Test} \\
\toprule
AOKVQA & 17056(13645 + 3411) & 1145 \\
PathVQA &19653(15723 + 3930)&6259 \\
VizWiz & 20523(16418 + 4105)& 4319 \\
\toprule
\end{tabular}
\label{tab:calib}
\end{table}

\subsubsection{Model Information}
We evaluated our method on four visual language models in our experiments; details of their layers, hidden sizes, and logit sizes are provided in Table~\ref{tab:models}, giving a brief overview of the sizes of the input vectors used to train the learnable scoring functions. Further, for inference, we use the same prompt for all models: 
\textit{\textless image \textgreater question, please provide a single word or short sentence answer. ASSISTANT:}
\begin{table}[h!]
\centering
\small
\setlength{\tabcolsep}{6pt} 
\caption{Architecture details of the pretrained visual language models used in our experiments.}
\begin{tabular}{c|c|c|c}
\toprule
\makecell{\textbf{Model} \\ \textbf{Name}} & \makecell{\textbf{Number of} \\ \textbf{Layers}} & \makecell{\textbf{Hidden Dim.} \\ \textbf{Size}} & \makecell{\textbf{Vocabulary} \\ \textbf{Size}} \\
\toprule
LLaVa-7B       & 33 & 4096    & 32,000   \\
LLaVa-13B      & 41 & 5120    & 32,000   \\
Instruct-BLIP  & 25 & 2048    & 32,128   \\
Qwen-VL        & 33 & 4096    & 151,936  \\
\toprule
\end{tabular}
\label{tab:models}
\end{table}


\subsubsection{Hyper-Parameters}
\begin{figure*}[!t]
    \centering
    \begin{subfigure}[t]{0.31\textwidth}
        \includegraphics[width=\linewidth]{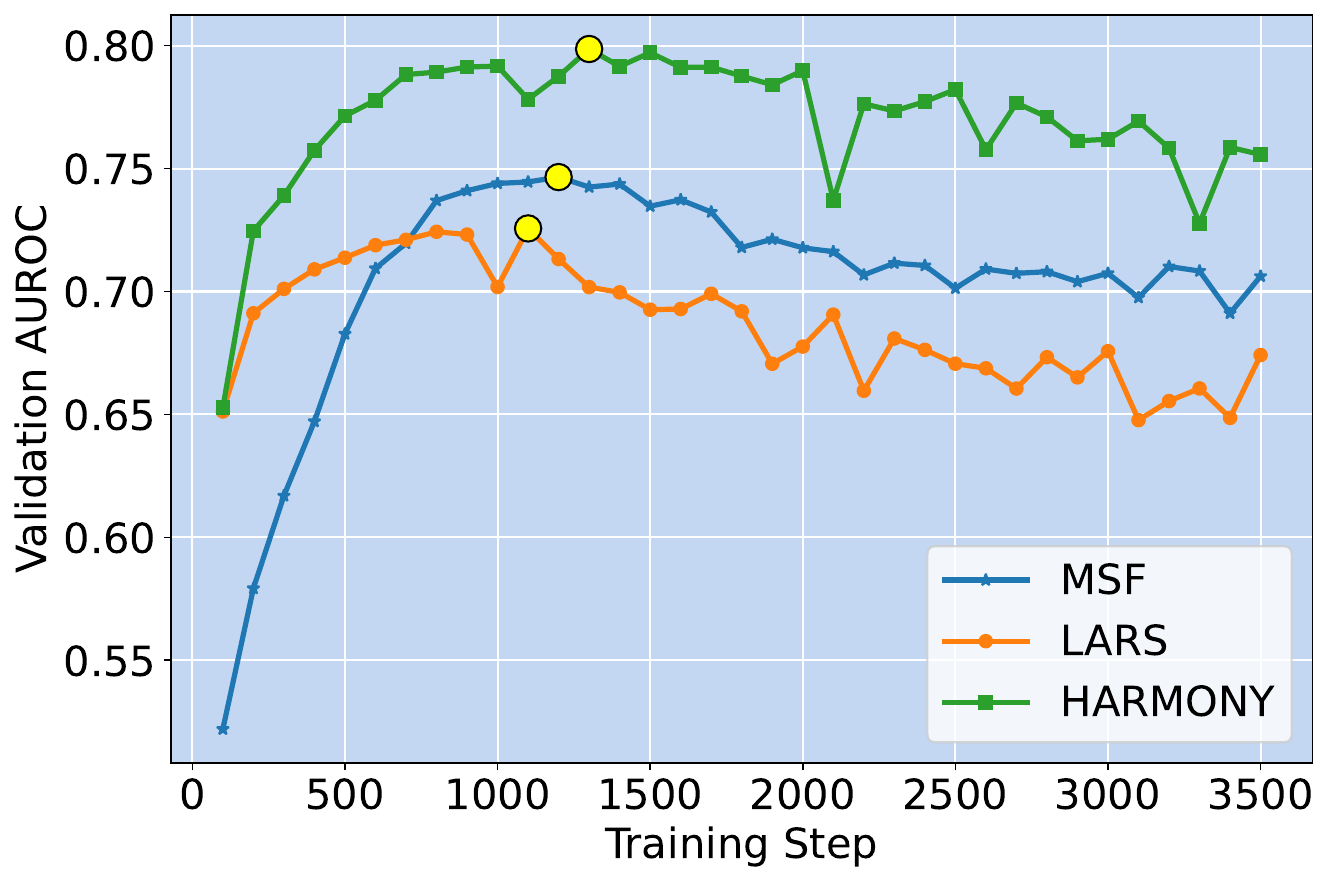}
        \caption{AOKVQA - Validation AUROC}
        \label{fig:aokvqa_val_auroc}
    \end{subfigure}%
        \begin{subfigure}[t]{0.31\textwidth}
        \includegraphics[width=\linewidth]{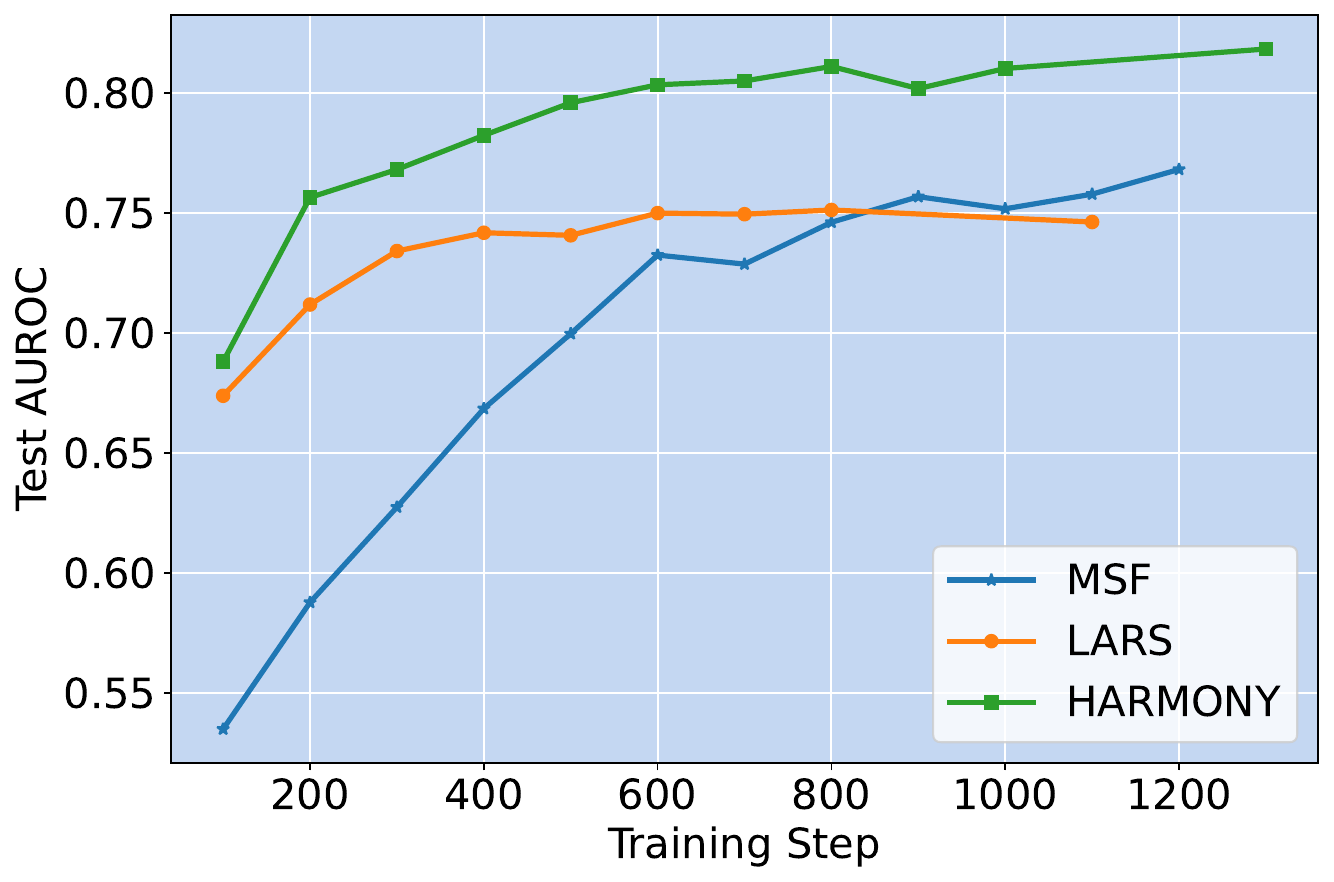}
        \caption{AOKVQA - Test AUROC}
        \label{fig:aokvqa_test_auroc}
    \end{subfigure}
    \hspace{0.02\textwidth}%
        \begin{subfigure}[t]{0.31\textwidth}
        \includegraphics[width=\linewidth]{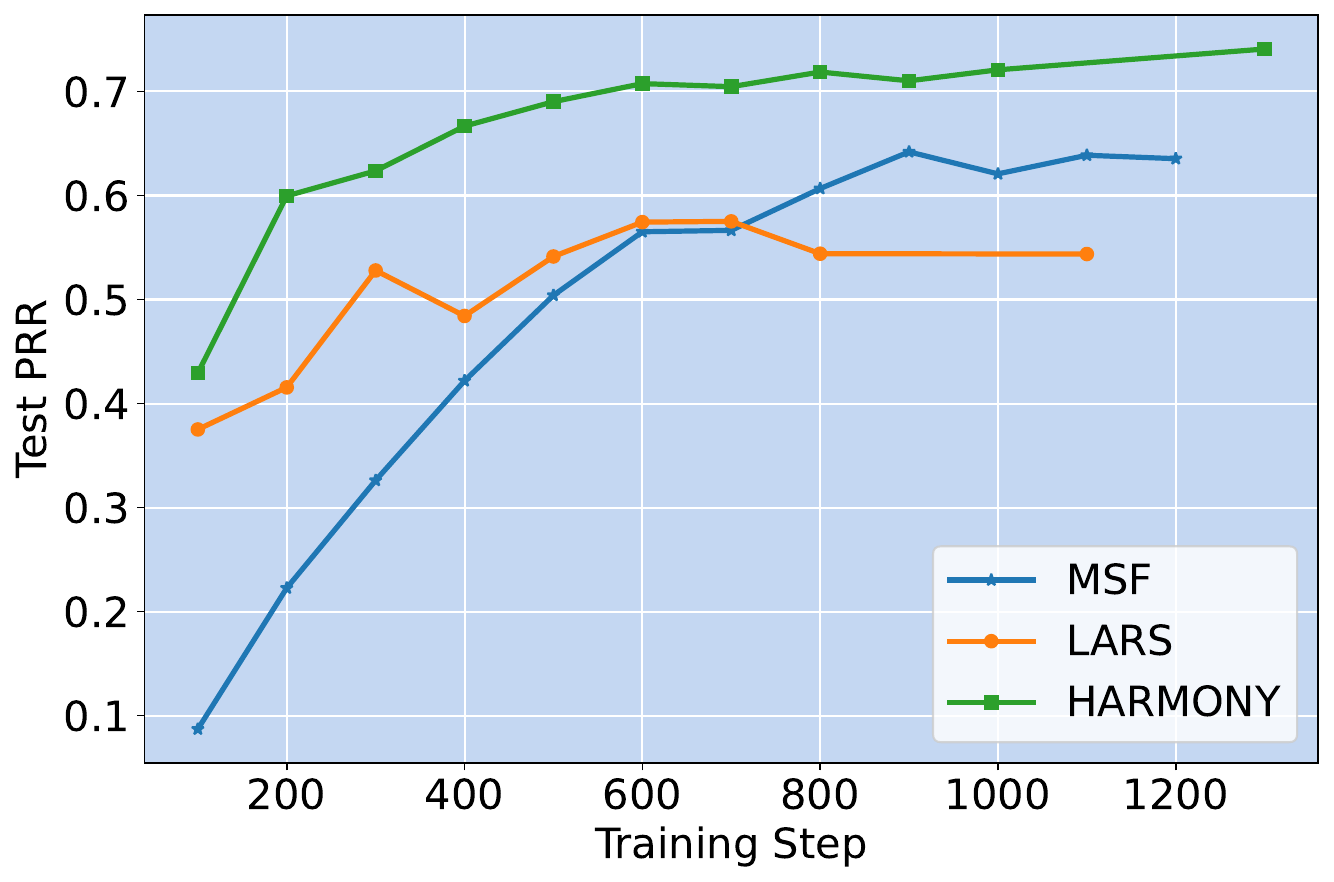}
        \caption{AOKVQA - Test PRR}
        \label{fig:aokvqa_test_prr}
    \end{subfigure}%
    \vspace{0.3cm} 
    \hspace{0.02\textwidth}%
    \begin{subfigure}[t]{0.31\textwidth}
        \includegraphics[width=\linewidth]{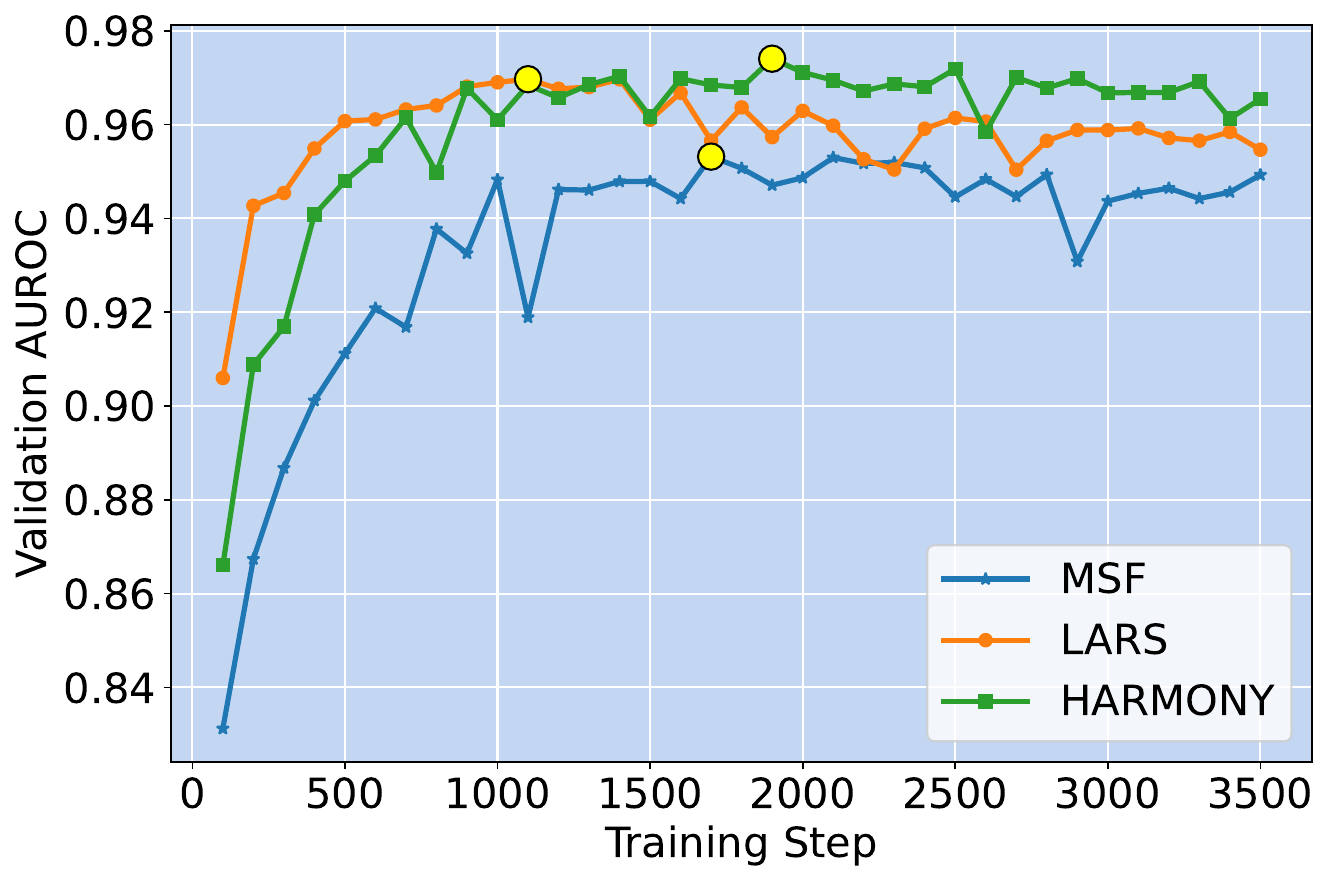}
        \caption{PathVQA - Validation AUROC}
        \label{fig:pathvqa_val_auroc}
    \end{subfigure}%
    \begin{subfigure}[t]{0.31\textwidth}
        \includegraphics[width=\linewidth]{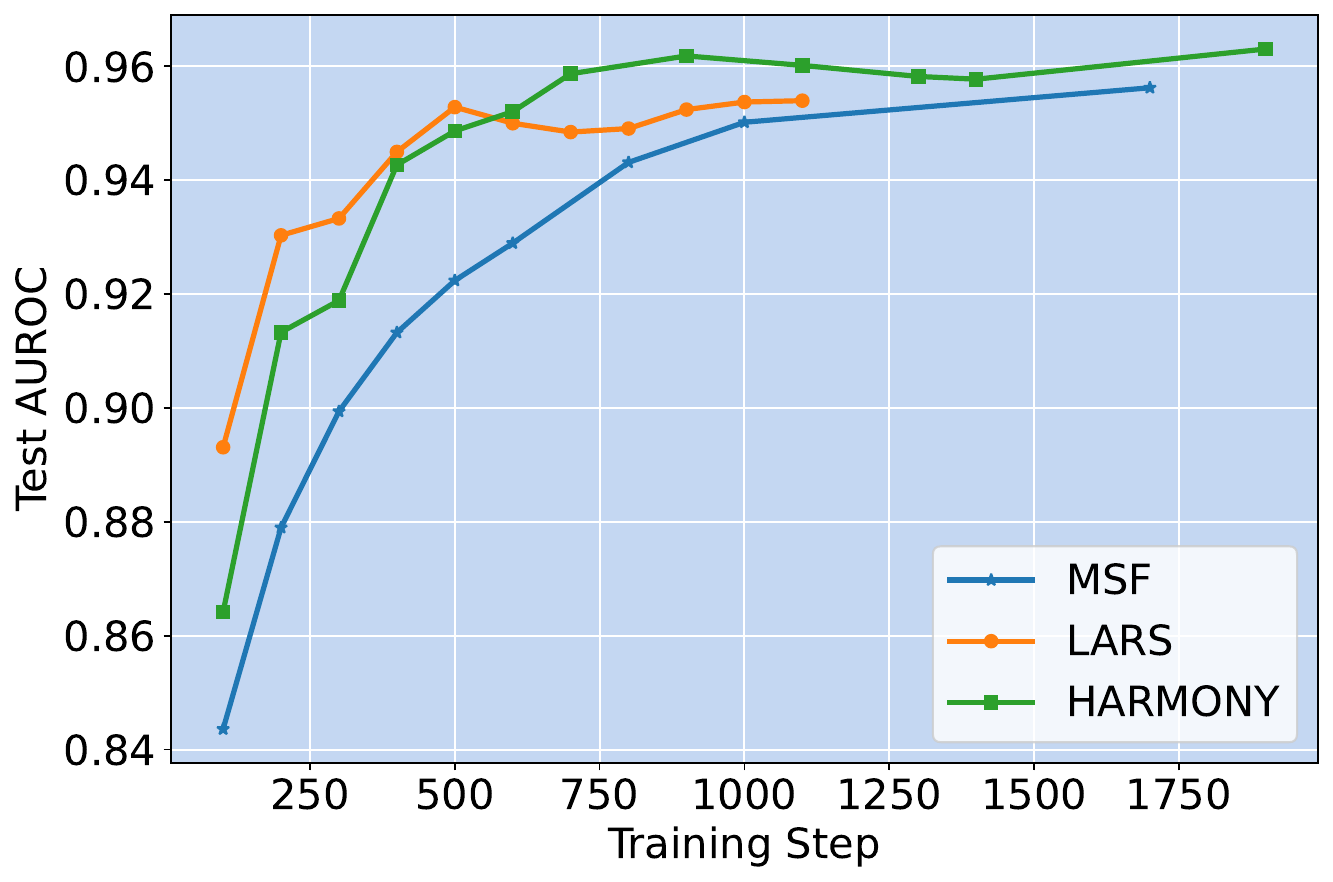}
        \caption{PathVQA - Test AUROC}
        \label{fig:pathvqa_test_auroc}
    \end{subfigure}%
    \hspace{0.02\textwidth}%
    \begin{subfigure}[t]{0.31\textwidth}
        \includegraphics[width=\linewidth]{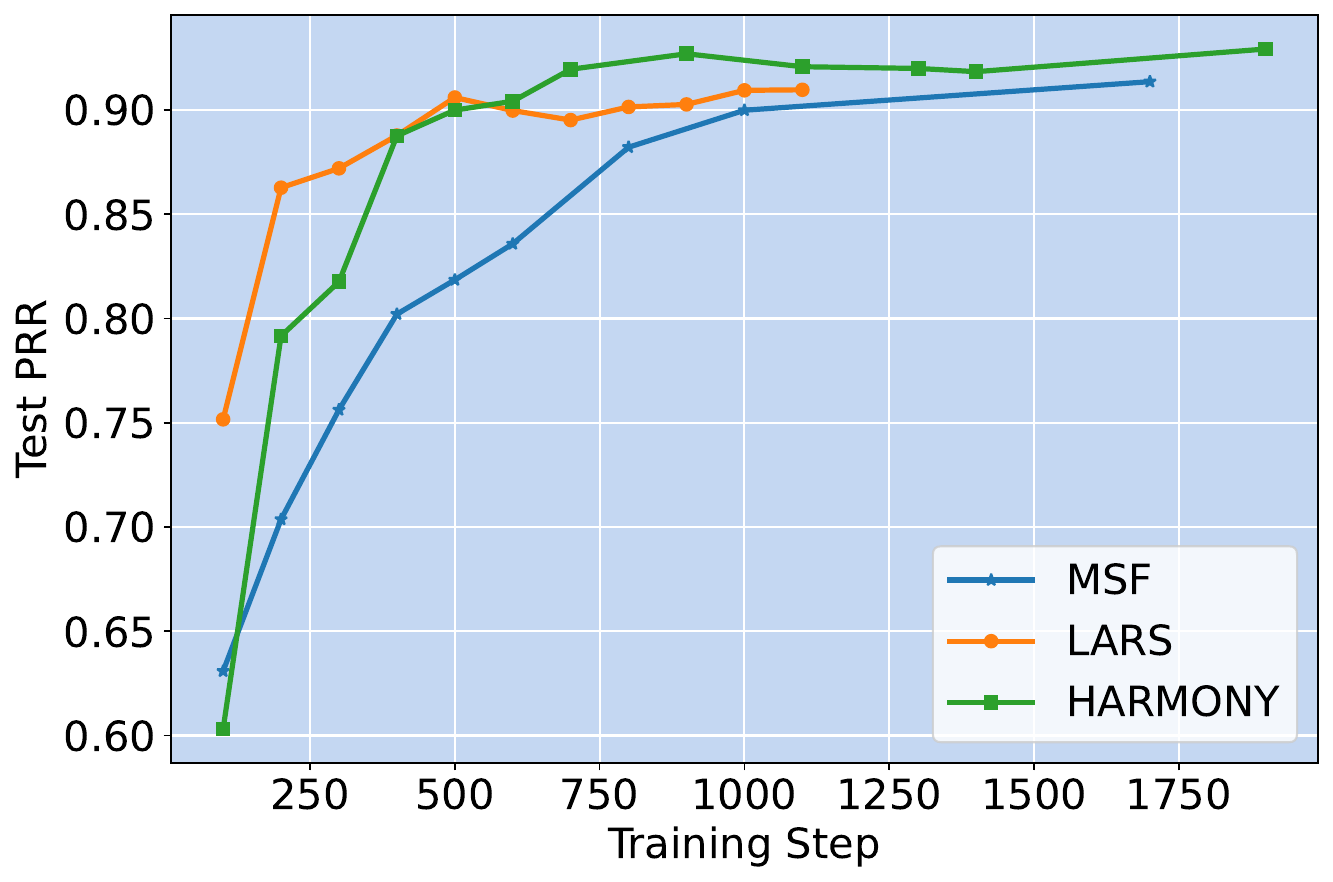}
        \caption{PathVQA - Test PRR}
        \label{fig:pathvqa_test_prr}
    \end{subfigure}
    \caption{Comparison of learnable scoring functions, MSF, LARS and HARMONY during training on the AOKVQA and PathVQA datasets using the Qwen-VL model. Validation AUROC is used to select the best checkpoint (highlighted with a yellow circle), and the corresponding Test AUROC and PRR scores are reported in the proceeding plots and Table \ref{UE_comparison_}.}
    \label{fig:training_runs}
\end{figure*}

We maintain two sets of data; calibration data and test data. Calibration data is further split into 80\% and 20\% split into training and validation data, respectively. For each model-dataset training, and every trainable scoring method, we perform hyper-parameter tuning of learning rate over \{ 5e-4, 5e-5, 5e-6\}. We found 5e-5 lr works best for most experiments. We use the AUROC metric as our best model checkpoint selection criterion. We also show this in Figure \ref{fig:training_runs} where we record AUROC and PRR on test set if the validation AUROC is >= the previous best. Finally, the best validation auroc step has been plotted by a yellow circle. For the best checkpoint, we report AUROC and PRR on the test set. We also use that checkpoint to report Coverage at risk $10\%$ and $20\%$, and effective reliability as reported in Table \ref{select_pred}. Further, for all the trainable methods, we use 20 epochs, and used early stopping; i.e, if validation auroc does not improve for 1K training steps, we stop the training. For MSF implementation, we follow the MLP architecture details from the official implementation of MSF (specifically, VisualBERT architecture experiments). We also keep other parameters such as optimizer (AdamW), learning rate scheduler (Warmup Cosine Scheduler) and batch size also the same. We use the same optimizer and learning rate for our method and LARS as well. Further, we use batch size of 32 for LARS and HARMONY. Since we are working on open-ended generations which can be of arbitrary length, therefore, we use zero padding to keep hidden representations of same length that is 128 for all the methods. For probability split in LARS, and our method, we use a bin count of 8.

\subsubsection{Additional Results on Medical Domain}
In Table \ref{medical}, we also report the UE performance on the medical imaging domain dataset, PathVQA with Qwen-VL model. For this dataset and model as well, we find that the output probability based scoring functions perform significantly lower than the learnable functions as shown in \ref{medical}.  Further, HARMONY achieves state-of-the-art performance indicating its effective performance in medical domain.
\begin{table}[h]
\caption{UE Performance on PathVQA}
\centering
\renewcommand{\arraystretch}{1}
\setlength{\tabcolsep}{10pt}
\resizebox{0.5\textwidth}{!}{
\begin{tabular}{ cl c c }
\toprule
& & \multicolumn{2}{c}{\textbf{Qwen-VL}}   \\
 
 \midrule
& UE Method & AUROC(\%)  & PRR(\%)  \\
\midrule[\heavyrulewidth]
&Length-Normalized Confidence  &  71.18&26.74 \\
&First Token Confidence   & 71.64&26.65\\
&Self-Eval Confidence         & 71.91&30.57\\
\cmidrule{2-4}
&Entropy       & 76.03&45.86\\
&Semantic Entropy      & 63.72&15.42\\
&Cluster Entropy      & 61.79&10.97\\
\cmidrule{2-4}
& MSF  &91.36&95.62\\
&LARS & 91.99 & 95.82\\
&HARMONY [Ours] & \textbf{92.92} & \textbf{96.30}\\
&&\textbf{(+0.93)}&\textbf{(+0.68)}\\

\midrule[\heavyrulewidth]

\end{tabular}
}
\label{medical}
\end{table}
\vspace{-20pt}
\subsubsection{Computation}
We perform all inference generations and evaluations on NVIDIA A100 40GB GPUs. For all inference generations and evaluations including black-box baselines, such as Semantic Entropy, we used two GPUs. Generation and labeling of the calibration data for one dataset containing ~13k-16k samples and one model with LAVE labeling \cite{manas2024improving} takes approximately 8-9 GPU-hours. Training a scoring function required at most two hours per run.

\end{document}